# Ontology Driven Disease Incidence Detection on Twitter

**Mark Abraham Magumba**

Department of Information Systems, School of Computing and Informatics Technology

College of Computing and Information Sciences, Makerere University, Uganda

magumbamark@hotmail.com

**Peter Nabende**

Department of Information Systems, School of Computing and Informatics Technology

College of Computing and Information Sciences, Makerere University, Uganda

peter.nabende@gmail.com

**Abstract**

*In this work we address the issue of generic automated disease incidence monitoring on twitter. We employ an ontology of disease related concepts and use it to obtain a conceptual representation of tweets. Unlike previous key word based systems and topic modeling approaches, our ontological approach allows us to apply more stringent criteria for determining which messages are relevant such as spatial and temporal characteristics whilst giving a stronger guarantee that the resulting models will perform well on new data that may be lexically divergent. We achieve this by training learners on concepts rather than individual words. For training we use a dataset containing mentions of influenza and Listeria and use the learned models to classify datasets containing mentions of an arbitrary selection of other diseases. We show that our ontological approach achieves good performance on this task using a variety of Natural Language Processing Techniques. We also show that word vectors can be learned directly from our concepts to achieve even better results.*

## 1.0. Introduction:

Unstructured data sources including social networking platforms such as Twitter and Facebook are increasingly being employed as data sources for disease surveillance [Schmidt 2012]. User data such as messages and user searches from such platforms have been used for both singular disease surveillance for example Google Flu Trends and Google Dengue trends for influenza and Dengue Fever [Google.com 2016] respectively and multiple disease surveillance for example by HealthMap.org[1]. These systems generally rely on a set of disease-related key words that filter a stream of documents to retain only those that are relevant. In this context, documents take on the form of messages posted by users; for example, in Twitter, documents are "tweets" which are short messages (limited to 140 characters in length) whereas, in Facebook, documents may comprise "wall posts", "comments", page feeds to mention but a few. Although huge amounts of textual data are generated by users of the Web-based social networking platforms, most of the textual data present some challenges for developing classification models for disease surveillance. For example, data sparseness is a major challenge when using Twitter data whose tweets have deliberate misspellings, made up words, non-standard grammar, and high lexical diversity [Jarvis 2013; Eisenstein 2013].

---

[1] http://www.healthmap.org/en/

In this paper, we propose an ontological approach for deriving features for classification models that can generally detect disease incidence in tweets and possibly other unstructured data sources. We require these models because these internet based surveillance methods typically count the volume of messages about a given disease topic as an indicator of actual disease activity. Generally speaking some positive correlation is assumed between the volume of messages about a given disease and its activity at a given time. However, in many cases this assumption is too weak as the volume of disease related messages can be influenced by panic and other factors as noted by Lampos and Cristianini [2010]. Therefore it is important to incorporate the semantic orientation of tweets as in many cases many messages about a given disease may actually mention it in a non-incidence related context or one that is spatio-temporally irrelevant. For instance, "I remember when the Challenger went down, I was home sick with the flu!" is an actual reference to an incidence of the flu but the Challenger disaster occurred in 1986 therefore it would be incorrect to count this mention for an outbreak in 2016.

In our approach we employ an ontology of concepts to overcome the problems of Natural Language Processing on Twitter. Our approach similar in many ways to the one applied by Collier et al [2008] in their BioCaster[2] system for detecting and tracking infectious disease outbreaks using RSS newsfeeds from ProMED-mail[3] service which is a global reporting system for outbreaks of emerging diseases and toxins. We have three main objectives: Firstly to create a catalog of words used to communicate illness, secondly to organize this catalogue into an ontology of concepts where concepts represent words that are related in meaning and finally to build classification models using our concepts as features. The classification task is to distinguish between relevant and irrelevant messages given a data set of tweets about a particular ailment.

## 2.0. Related Work:

The use of online unstructured data sources for disease surveillance constitutes a very active area of research. The general approach involves the use of a list of keywords to filter a stream of documents. For some systems like BioCaster [Collier et al, 2008] these key terms have been arranged into ontologies. Ontologies are used for explicit knowledge representations for inference support. In the domain of medicine and epidemiology, several ontologies have been created for the purpose of creating standardized vocabularies such as International Health Terminology Standards Organization's SNOMED-CT (Systematic Nomenclature of Medicine – Clinical terms), the Syndromic Surveillance Ontology (SSO) [Okhmatovskaia et al 2009], the Dictionary of Epidemiology [Porta 2008], the BioCaster Ontology [Collier et al 2007], the OBO ontologies such as DOID ontology for diseases [Osborne et al], OBO TRANS for disease transmission and SYMP ontology for symptoms [Scriml et al 2010] and OBO VO for vaccines [Yang et al 2011], the Epidemiology Ontology (EPO) [Pesquita et al 2014]. These ontologies have been developed with different foci and audiences in mind. For the most part the goal is to arrive at a standardized domain terminology for instance for SNOWMED-CT which is an attempt to harmonize divergent medical terminology and the European Epidemiology Ontology which does the same for select diseases. For the purpose of online disease surveillance, the BioCaster ontology [Collier et al 2008] and BioStorm (Biological Spatio-Temporal Outbreak Reasoning Module) [Buckeridge et al 2008] are the primary examples.

---

[2] https://sites.google.com/site/nhcollier/projects/biocaster
[3] http://www.promedmail.org/

In our case our focus is to organize the words that people use to communicate illness with an aim of facilitating training of general purpose classifiers for disease detection with low variance on divergent data sets. As Twitter users are generally less technical we give less focus to technical terminology such as names of specific diseases, pharmacological terminology or strains of disease pathogens. Regarding the use of Twitter for disease surveillance, there already exists an impressive body of work including several ongoing projects. In terms of text processing, these fall into two broad categories: that is, rule-based systems and machine learning systems. Rule-based systems use static approaches such as the inclusion of particular keywords such as the work by Lee et al [2013]. Machine Learning Systems come in two flavors: supervised systems which train a learner using a set of labeled examples such as the Crowdbreaks project[4] and unsupervised approaches such as Latent Semantic Analysis (LSA), Latent Dirichlet Allocation (LDA) and the LDA based ATAM (Ailment Topic Aspect Model) [Paul and Dredze 2014] which employ some theory such as the Distributional Hypothesis [Harris 1954] to arrive at some latent features. In addition, there are projects that combine both methods such as HealthTweets.org [Dredze et al, 2014].

Where supervised methods are employed at some point, obtaining labeled datasets is a challenge and different strategies have been tried by different projects; for instance Paul and Dredze [2014] use an SVM (Support Vector Machine) classifier during corpus generation for their ATAM model and rely on Amazon.com's Mechanical Turk (Mturk) tool [Callison-Burch and Dredze 2010]. Crowdbreaks relies on crowd sourced annotation from site visitors. However, it is a huge undertaking to create a statistically representative amount of annotated data which naturally hinders the applicability of supervised models to a general purpose, online application.

### 3.0. A Concept Ontology of Human Disease Language

We start out with the observation that people tend to use the similar words when communicating illness, furthermore there are certain words that are specific to certain diseases or groups of illnesses. These lexical differences originate from differences in diseases such as the site of infection. For instance in a collection of tweets describing respiratory infections the words *cough, sneeze and sniffle* are likely appear frequently whereas in a collection of tweets describing dermatological ailments the words *itch, scratch, burning and eczema* may be more frequent. If a word level classifier were built using a dataset that has a high proportion of tweets about respiratory illnesses, it may not encounter any tweet containing the word *eczema* even though it might be a strong predictor for skin disease. However, it is easy to see that *cough, sneeze, eczema* are part of a larger group of concepts, that is, they are all diseases symptoms of diseases. We can even go further and recognize that these words are part of a larger concept which is morbidity or being sick and hence bundle all sickness related words into a single concept.

Furthermore, a word level model may achieve high classification performance on a given data set but good performance on new data requires that the training data was representative. However, on a medium like twitter there is such a huge lexical diversity that many words will be encountered just a handful of times even in a large data set meaning it will never learn certain relationships. This also means that many of the relationships learned by a learner may be spurious for instance in our own experiments when using unigram bag of words on the flu dataset which contains tweets about flu and Listeria we find that the word *game* is a strong positive predictor whereas its plural *games* is a strong negative predictor. This is because of phrases like "I am going to have to bring my flu game" which refers to indulging in sports whilst sick with the flu in homage to a remarkable feat by the legendary Micheal Jordan when he led the Chicago Bulls to victory against Utah Jazz in the 1997 NBA final whilst stricken with the flu. "Flu games" on the other hand, is a reference to a brand of sporting footwear in homage to the same game. However, in a dataset containing tweets about norovirus or even tweets about the flu but from a geographic location outside North America linguistic differences may make such an inference invalid.

---

[4] http://www.crowdbreaks.com/

However, we observe that there are some higher concepts that regardless of lexicon that are used to communicate illnesses. As an example, as earlier observed, words that refer to sickness or morbidity such as symptoms. By identifying such concepts and representing messages in terms of these concepts we can train a learner to learn relationships between concepts as opposed to relationships between individual words. In theory such a learner should generalize better as we expect its performance on lexically divergent datasets to be less variant therefore making a generic classifier more feasible.

We conceptualize each message as a concatenation of references to some higher conceptual objects. For the purpose of communicating disease incidence we model an ontology representing a hierarchy of concepts that describe disease incidence. These concepts are of two broad categories, those that directly describe disease incidence such as references to disease causing organisms such as bacteria and general linguistic terminology like words that imply negation and references to temporality such as words like "now" and "then" and descriptions of space and time. As depicted in figure 1, At the top of the hierarchy everything is conceptualized as an object, there are two types of objects that is real objects and abstract objects. Real objects refer to tangible things and abstract objects refer to concepts like time, negation and events. Real objects are either Living or inanimate.

Living things comprise all the key biological actors such as hosts (the organism that suffers disease which in this case is a person), pathogens (disease causing organisms) and vectors (disease spreading organisms). Inanimate objects are of two major classes namely environments and substances.

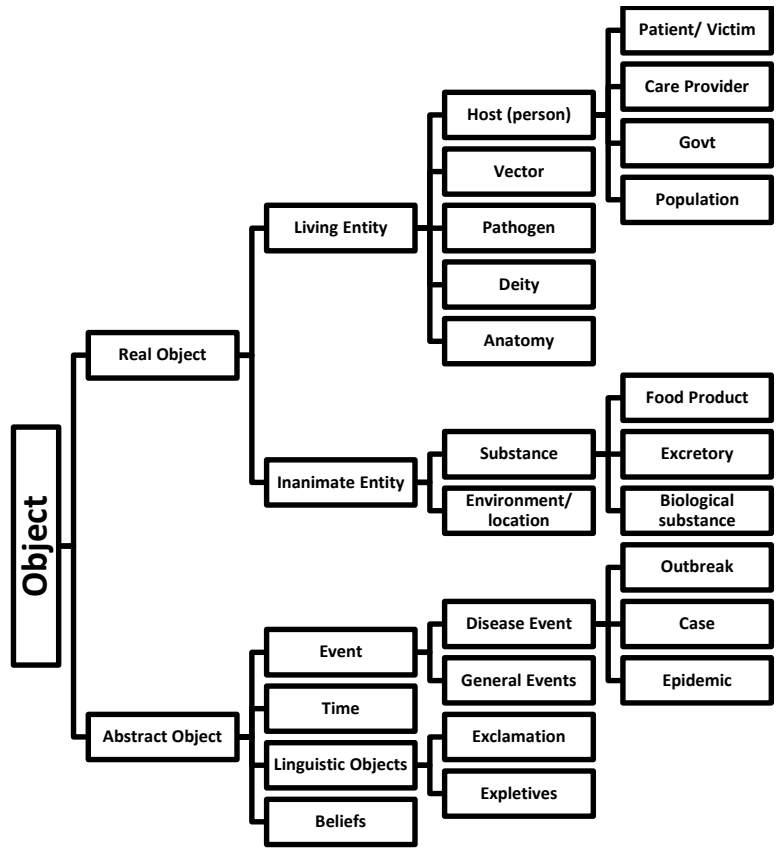

*Figure 1: Partial Class Hierarchy for Human Disease Language Concept Ontology*

There are three additional types of concepts namely relationships and properties and actions. Relationships describe interactions between concept classes in the object hierarchy and correspond to OWL (Web Ontology Language) object properties whereas properties describe object and relationship attributes and correspond to OWL data properties.

The concept hierarchy allows for additional semantics including sub classing and inheritance. For instance figure 2 depicts sub classes for the People object excluding anatomy and population sub classes along with relationships and classes. Inheritance means that all definitions made in the top level object are inherited by the child classes in this case care provider, patient and government. In addition there is support for some polymorphic behavior meaning that even though inheritance allows for one time definition shared behavior in the hierarchy it is not necessarily implemented uniformly. For instance all objects have a "Magnitude" property that allows for quantification but this is invoked differently depending on the object for instance it could be a discrete quantity or quantification by degree.

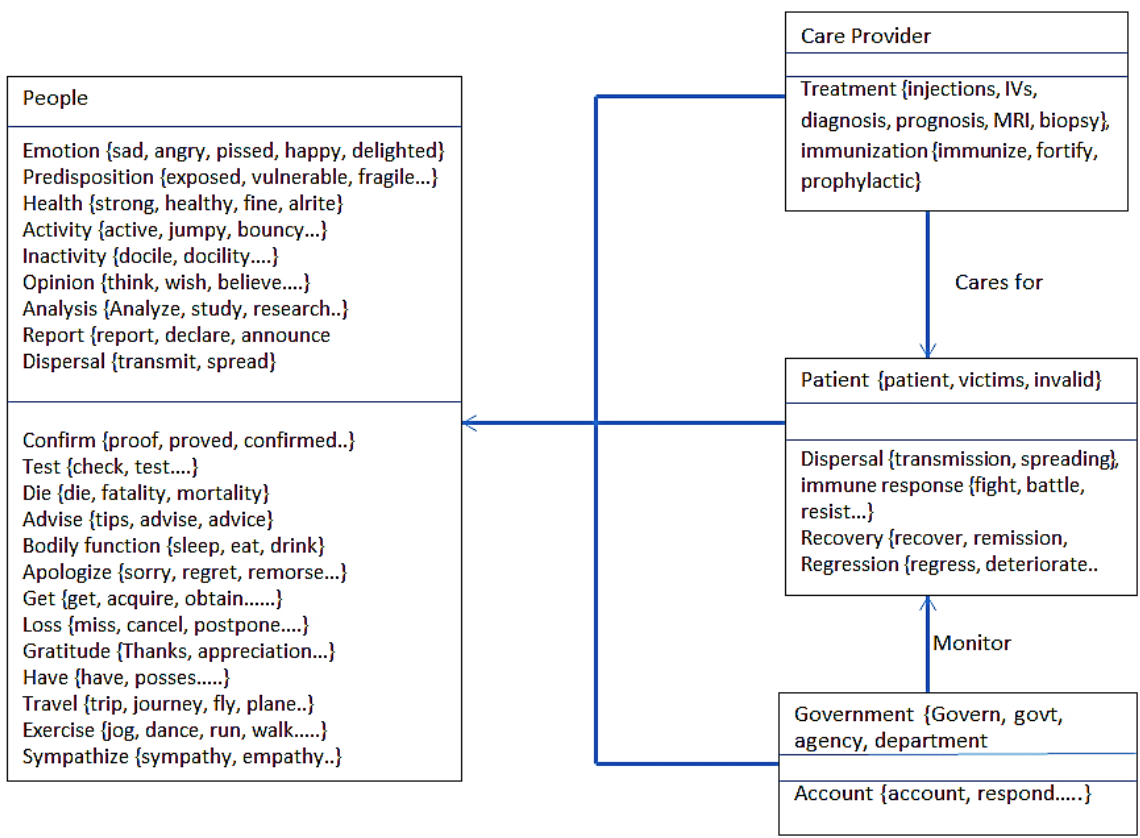

*Figure 2: People Object sub hierarchy and concept dictionaries*

### 3.1. Concept Dictionaries:

Finally, for each conceptual object be it a relationship, object or action there is a corresponding concept dictionary. In figure 2 these are indicated by the curly brackets. We also use a UML (Unified Modeling language) class diagram instead of the more appropriate RDF schema diagram because it is more compact. However, the full ontology is publically available in OWL/XML format at our Github repository[5]. In this case object properties in which the domain and range are the same class have been represented as UML class methods. The concept dictionary is simply a list of words related to the concept. Roughly speaking the members of a given concept dictionary are related by three relationships namely synonymy, hyponymy/hypernymy and meronymy. Synonymy refers to semantically equivalent words such as "skin" and "dermis", hypernymy is refers to increasing generalization for instance "location" is a hypernym of "house", hyponymy is the inverse of hypernymy as in "house" would be a hyponym of "location". Generally speaking immediate hyponyms and hypernyms are grouped together with concept synonyms. Meronyms are similarly grouped together, meronymy implies a part of relationship for instance "arm" is a meronym of "body".

However the principle relationship between members of a concept is thematic for instance the "treatment" concept bundles together treatment nouns like "stethoscope" and treatment verbs like

---

[5] https://github.com/MarkMagumba/Twitter-Disease-incidence-Description-Language-Ontology

"diagnosed". Also as an implication in some cases even antonyms are bundled together for instance the two word "large" and "small" are both bundled into the "extent" concept dictionary.

### 3.2. Temporality:

Temporality is important for determination of relevance for instance the message, "I have the flu!" is relevant whilst "I have had the flu only once in my life" though referring to an actual case of the flu is not relevant. Time is dealt with by the time object which is an abstract object. Time is separated into two periods namely the window and other temporal references. The window refers to references to disease events that fall approximately within the communicable period which is the time period within which a disease is likely to be spread from one host to another. In this time it can be said there is an active case of a disease or reference is made to recent case of a disease with a maximum span of a few weeks. The window includes words like "today" and "yesterday" whereas other temporal references include words like "year" and "decades" where it is extremely unlikely that reference is being made to an ongoing case of a disease. Furthermore we take into account different facets of event history in the form of aspectual predication similarly to TimeML [Sauri et al 2005] which defines five event states shown here with examples (keywords in caps)

1. Initiation (the event is beginning e.g. "Flu season has BEGUN"),
2. Re-initiation (an event that stops then starts again e.g. "I am falling sick AGAIN!"),
3. Continuation (the event is continuing e.g. "I am STILL down with the flu")
4. Termination (An event that would otherwise continue is stopped e.g. "I stopped my pain medication")
5. Culmination (A process coming to conclusion e.g. "I completed my therapy!")

In our work we make no distinction between initiation and re-initiation and combine them into a single "Commencement" concept. In the case of general events we do not make a distinction between culmination and termination. However in the case of disease events, we take care of culmination via two specialized event types namely Mortality (A patient dying from a disease) and Recovery (A patient going from being sick to being healthy. In addition as seen in figure 3 we define certain trend events which describe the direction of change of state subject to some interpretation and these are of three general types namely increment (A net positive change) ,stasis (No net change) and reduction (A net negative change).

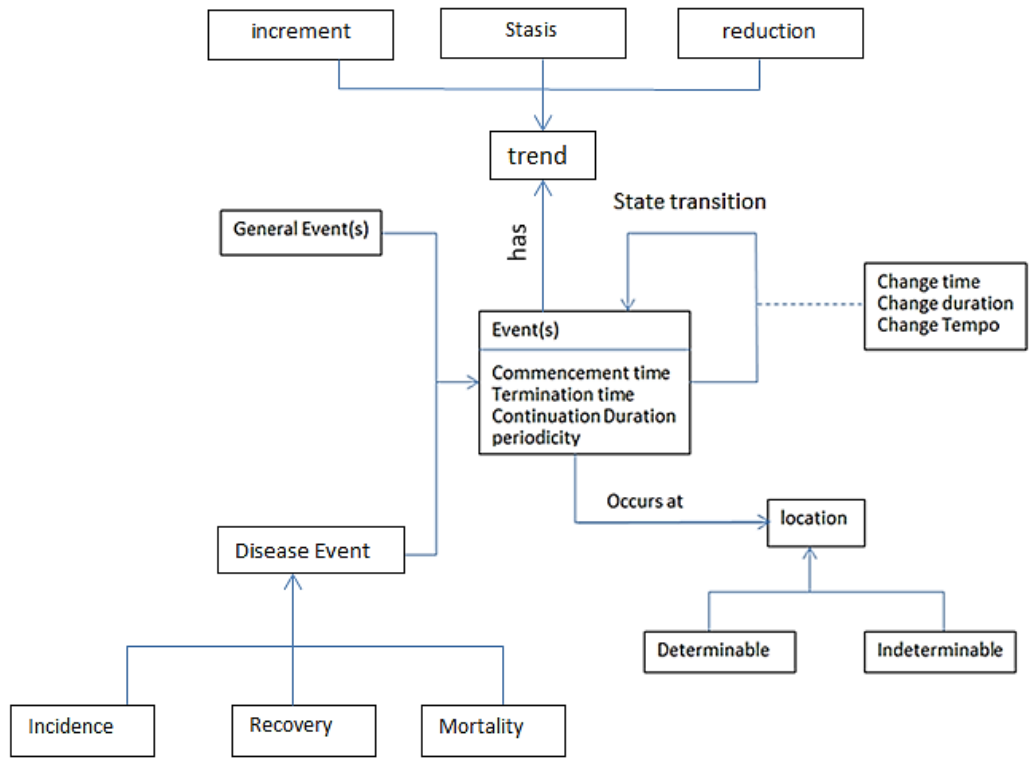

*Figure 3: Event Handling Overview in Disease Language Ontology*

Events also have duration and periodicity. Durations can be expressed in time lengths (e.g. "days", "weeks" as in the window or in relative terms such as "until", "since"), periodicity refers to frequency for instance words such as "never" and "always". Finally temporality is dealt with by tensing. Generally we take advantage of English structure by relying on agreement of tense. For instance we do not make a distinction between the two words "overcoming" and "overcame" as these are simply returned as references to recovery. This distinction is only made for a few common words such as "have" and the verb to be, where mortality occurs in the same sentence as "have" it can be concluded that mortality must exist in the past tense. Tensing is further strengthened by Part of Speech Enriched Padding discussed in section 4.2.

### 3.3. Locality:
As depicted in figure 3, events also have locality. Locality is indicated by words such as *home, school, state* and in other cases is easily inferred particularly when users speak in the first person as even without an explicit reference to locality it is easily deducible that the user's location is also the disease event location . It is also denoted by prepositions like *near, behind and adjacent.* However in many cases the location is indeterminable. For a tweet to be classified as relevant both time and locality must be determinable.

### 3.4. Magnitude
Two types of magnitudes are catered for that is discrete quantities and extents. Quantities refer to countable magnitudes like number of victims and inexact quantization for instance when quantities are conveyed vaguely via words like "many" and "several". We do not perform any quantification and only keep track of quantification words for two particular purposes firstly where quantities are used as indicators of persons for instance in the sentence "HUNDREDS contract deadly norovirus" and secondly where quantities are used for negation for instance "NONE of my workmates has zika virus". As with temporality, where quantities are used to represent persons we divide them by scale into two groups. The two groups are smaller quantities up to thousands and quantities larger than thousands. This is from the observation that disease incidence is reported in small numbers of up to a few thousand. Statements that refer to disease incidence in larger quantities such as "Millions die from Malaria in Africa every year" and "Billions lost due to norovirus annually" are generally informational and serve no purpose in a disease incidence monitoring sense. Extents on the other hand refer to quantification by degree via words like *large, mild, complete* and *partial*.

### 3.5. General Order

All objects are also subject to order which can be spatial, temporal or otherwise and is conveyed by temporal connectives such as *before, after, while* and positional adjectives such as *fast, last, second, next and preceding.*

### 3.6. Linguistic Elements

Linguistic elements include additional concepts such as exclamation, expletives (e.g. "bitch", "Damn"), prepositions which are further broken down for our purposes. For instance even though the two words "on" and "off" are grammatically both prepositions they are semantically opposite, the same applies to conjunctives such as "but" and "and" which have totally different implications, and determiners (this, that, those) which may also carry different contextual implications. Additional concepts include negation which includes words such as "not", "neither" and "none".

### 3.7. Semantic Ambiguity

In some cases there is semantic ambiguity for instance the word "fall" can imply a kind of motion as in "I am falling down" and also getting sick for instance "I am falling sick". It would be inaccurate to bundle it with other motion verbs like "jump" as that would be to tell the learner that they always imply similar connotations. Particularly in the case of polysemy, words may present different meanings meaning they can legitimately belong to multiple concepts, in this situation the most semantically accurate approach is to create special single word concepts.

### 4.0. Feature Generation:

We transform each tweet into a vector of features as follows. Firstly we flatten out our ontology into a set of concepts. As stated in section 3.1 each concept is associated with a group of words or tokens referred to as the concept dictionary. Each concept is effectively a list of words and the full ontology is basically a list of lists. In this sense it is a heavily redacted English dictionary containing only words we consider to be of epidemiological relevance. We also avoid technical terminology like names of drugs and specific diseases and focus on words that would appear in neighboring contexts.

To obtain the feature vector we simply tokenize each tweet and for each token we do a dictionary look up in our flattened ontology. If the token exists in the ontology we simply replace it with the corresponding concept. Figure 4 below is a pseudo code summary of the approach. We refer to the resulting representation as the Concept Normal Form (CNF). The result is that we can represent any document using a fixed set of concepts. Effectively we have a fixed size lexicon regardless of the original contents of any document or tweet.

```
FEATURESETS = empty list
FOR example in EXAMPLES
    featureVector = empty list
    FOR token in TOKENIZE(example)
        CONCEPT = NULL
        FOR concept in CONCEPTS
            FOR word in DICTIONARY(concept)
                IF word == token
                    CONCEPT = concept
                    BREAK LOOP
                ELSE
                    Continue
        featureVector.APPEND(CONCEPT)
    FEATURESETS.APPEND(featureVector)
```

*Figure 4: Pseudo Code of the Procedure for Document Transformation into CNF*

As an example the sentence "I have never had the flu" is transformed into "SELF_REF HAVE FREQUENCY HAVE THE OOV". SELF_REF refers to "Self references" which is the concept class for terms that persons use to refer to themselves such as "I", "We" and "Us" used as an indicators of speaking in the first person, "HAVE" is the concept class for "have" or "had" which is a special concept class since the verb "to have" is conceptually ambiguous as it can legitimately indicate two senses that is falling sick or possession. The "FREQUENCY" terms refers to a reference to frequency concept which denotes temporal periodicity.

The "OOV" term at the end of the CNF representation stands for "Out of Vocabulary". To arrive at the concept representation we merely perform a simple list lookup; for each term we iterate through all the categories to see if it exists in any category if it does then that category label is returned, otherwise the "OOV" flag is returned. Using our object model we generate 1531 terms corresponding to 136 categories for our training data which has a vocabulary of about 59,000 words. Needless to say most words are out of vocabulary. Rather than completely ignore these we introduce a two-step categorization as described in section 4.1.

### 4.1. Part of Speech Enriched Padding

Instead of completely ignoring out of vocabulary tokens we replace them with their part of speech tag therefore our previous example, "I have never had the flu" becomes "SELF_REF HAVE FREQUENCY HAVE THE NN". For Part of Speech tagging we employ the Penn Tree Bank Part of Speech tags [Taylor et al 2003] in addition to some special tags for twitter specific phenomena such as "RT" for re-tweet, USR for users denoted by tokens beginning with "@", HT for "hashtag" denoted by tokens beginning with "#" and "URL" for http universal resource locators denoted by tokens beginning with "http". Figure 5 below depicts the transformations for the message "I have never had the flu!"

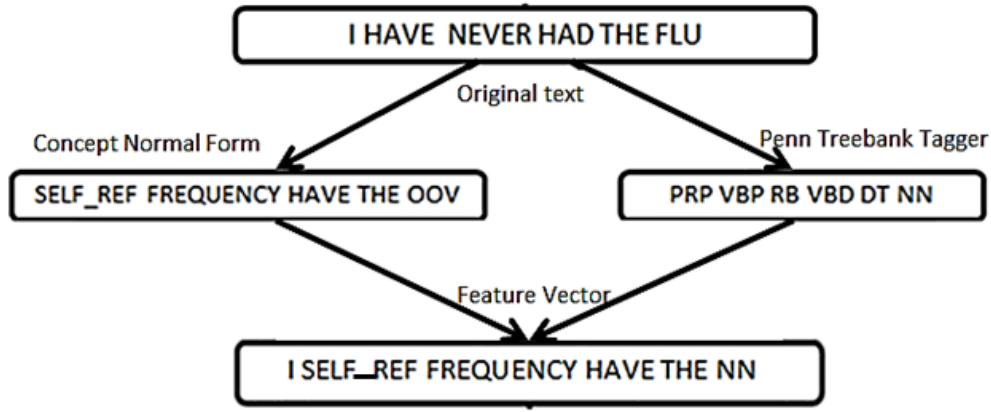

*Figure 5: Deriving Feature Vector from CNF and POS Tags*

The value of this step is that it introduces more robust tensing because the Penn tree Bank standard differentiates between tenses and as stated before we can take advantage of agreement of tense in English language. As a result of the huge lexical diversity of twitter in some cases it is more useful to cluster terms using their part of speech information. Furthermore, state of the art part of speech tagging has been shown to be at least 90% accurate. This means that even though many terms will be out of vocabulary we are still able to determine any word's grammatical category with a fair amount of certainty and the learning algorithm can potentially learn more for instance without this step, "I have never had the flu" and "We have often had dinner parties" are indistinguishable as they would have equivalent CNF representations. For part of speech tagging we employ the GATE[6] twitie tagger application [Cunningham et al 2002; Derczynski et al 2013] which is a variant of the log linear maximum entropy Stanford Part of Speech Tagger [Toutanova and Manning 2000; Toutanova et al 2003] trained on twitter data.

---

[6] General Architecture for Text Engineering

## 5.0. Methodology:

The following is a pipeline for generic disease incidence detection on Twitter using our ontological approach. Figure 6 below summarizes the steps

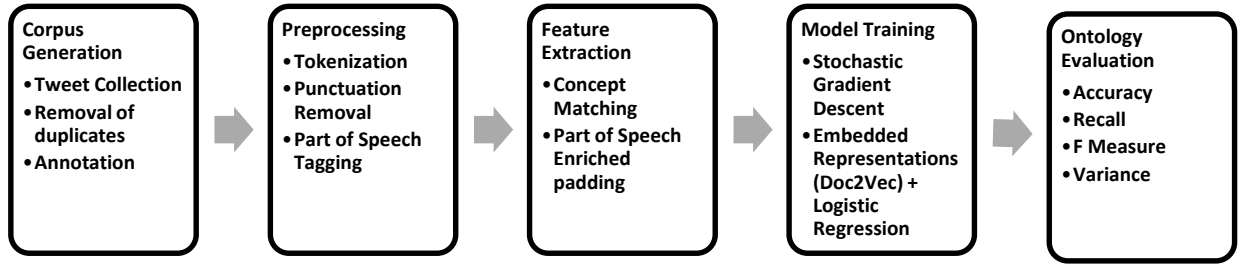

*Figure 6: The Disease Incidence Detection Pipeline*

### 5.1. Corpus Generation:

The pipeline is depicted in figure 6. The first step is the creation of the corpus. We obtain tweets from a basic twitter account using some specific keywords via a python script through Twitter's Streaming API. The tweets we download are those that are marked as public, this is the default security level and they are only marked private if expressly indicated by users. Next we partition the tweets into training and testing data sets. We created corpora for four diseases namely influenza + common cold + Listeria, stomach flu (gastro enteritis), norovirus, conjunctivitis 1 and conjunctivitis 2. There are two conjunctivitis data sets because for some tweets it is referred to as "pink eye" (conjunctivitis 2) and we were interested what impact such small linguistic differences may have on our models.

At this phase we eliminated duplicates as much as possible by removing any tweets with the "RT" tag (retweets) and manually reading through the tweets as many tweets were practically the same even though they did not have this tag differing only in minor details like the urls they contained. At the annotation phase we simply label each tweet as either positive or negative, where a positive tweet is a reference to an actual case of disease in the required time window which is basically the present time or the previous past to a period of a few weeks.

The influenza + common cold + Listeria data set comprises 13004 tweets of which 57.8% are positive and 43.2% are negative, The gastro enteritis data set comprises 646 tweets of which 84% are positive and 16% are negative, the noro virus data set comprises 1288 tweets of which 51% are positive and 49% are negative, conjunctivitis 1 comprises 656 tweets of which 55% are positive and 45% are negative and conjunctivitis 2 comprises 721 tweets of which 62.5% are positive and 37.5% are negative. The influenza + common cold + Listeria data set is used as the training data set and the other data sets as the evaluation data sets.

### 5.2. Preprocessing:

This involved transforming all tweets to lower case, removal of punctuations and part of speech tagging. For normalization and part of speech tagging we used GATE framework's Twitie tagger application after which we performed feature extraction using the method described in section 4.

### 5.3. Model Training:

We employed the following methods, first we did a Bag of Words (BOW) model using our original, untransformed dataset with a Stochastic Gradient Descent (SGD) classifier to obtain a baseline, then we extracted features using the procedure in section 4 and created a unigram Bag of Words model, a uningram + bigram model and a Doc2Vec + Logistic Regression classifier. For the Bag of Words models we used the SGD classifier with the same parameter settings.

Using Distributed Word Embeddings for text classification is a two-step process, first word vectors must be learned for each word in the vocabulary which is an unsupervised step then these vectors can be fed into a supervised learning method. The Doc2Vec algorithm is an extension of word2Vec first described by Mikolov [2013]. The word2vec algorithm takes a collection of documents and returns a vector (word embedding) for each word. The similarity between two words can be calculated from the cosine distance between their word embeddings. Generally vectors of words that are close in meaning are also close in value. It has also been shown that word2vec is capable of preserving non trivial semantic relationships between words for instance the vector for "Brother" – "Man" + "Woman" produce a result that is close to "Sister" (Wikipedia, 2016). Word2vec may employ one of two approaches, either skip gram or Continuous Bag of Words (CBOW). DocVec extends word2vec to entire documents. Roughly speaking, there are two ways in which this can be achieved. That is by computing the average of context word vectors or the concatenation of context vectors. In our case we use the former approach.

There are several parameters that need to be tuned to obtain good word vectors but some of the more important ones include the size, the window and whether or not to use negative sampling and the number of noise words to be drawn. The size refers to the dimensionality of the word vectors. Generally, the more dimensions the better the performance up to some point where additional dimensions do not significantly alter performance. The window specifies how many context words to use. For instance if the window size is set to $\beta$, then for the CBOW architecture if the current word is at index $i$, the task is to predict the most likely words at positions $i-1, i-2.....i-\beta$ and $i+1, i+2.....i+\beta$. For skip gram architecture the learning task is the inverse that is to predict the most likely ith word given the words at positions $i-1, i-2....i-\beta$ and $i+1, i+2.....i+\beta$. The corresponding architectures to CBOW and skip gram in DocVec are distributed memory and distributed bag of words. In addition we found that increasing the number of training epochs improved the results. We obtained our best results with 200 dimensions, with negative sampling with 8 noise words, a context window of 5, and with 20 epochs with distributed memory. For a deeper explanation of these parameters see (Rehurek, 2016).

When dealing with individual words determining model quality is straight forward for instance you would expect the vectors for "human" and "person" to be close but when dealing with coarse grained concepts as with our ontology we can't apply the same reasoning directly. This is because the concepts actually abstract several words as with the example given in section 3.1 of the treatment concept which bundles all treatment related words including treatment nouns and treatment verbs in different tenses. It becomes very difficult to determine for instance whether this concept is semantically closer to the immunity concept which bundles all words related to immunity or the morbidity concept which bundles all words that relate to the state of being sick such as "illness" and all symptoms we are presently able to list.

However since the final representation is a combination of concepts and Part of Speech tags for out of vocabulary words, we can instead use the Part of Speech tags where it is more straightforward to determine such similarity. For instance in a good model we would expect that the closest "concepts" to the NN (Noun, Singular) tag are NNS (Noun Plural) and NNP (Proper Noun Singular) and the most similar to JJ (Adjective) would be JJR (Adjective, Comparative) and JJS(Adjective, Superlative). In our model we find the closest concepts to NN are indeed NNP and NNS but for JJ we find NN and NNP to be closer than JJR and JJS which do not feature among the ten most similar concepts to JJ. This is not linguistically optimal but understandable. The word2vec algorithm relies on the distributional properties of individual tokens and adjectives occur in the same contexts as nouns and therefore would appear close in meaning to nouns. In addition as per our training data there are very few occurrences of JJS and JJR in comparison to JJ and NN. Nonetheless as we show in section 5.4, the learned vectors dramatically improve the results.

For Doc2Vec we employ our full corpus with the labels removed for the unsupervised portion. For the supervised learning portion we employ a logistic regression classifier. All code is written in python and we use the Scikit-learn package for all machine learning tasks [Pedregosa et al 2011] except the Doc2Vec algorithm for which we use the gensim package [Rehurek and Sojka 2010].

### 5.4. Model Evaluation:

The purpose of our ontology is to arrive at some symbolic representation of any document for the purpose of disease incidence monitoring. The resulting notation referred to here as the Concept Normal Form effectively compresses the vocabulary into a fixed length lexicon corresponding to a fixed set of conceptual objects presented in the ontology. To prove the quality and completeness of our representation we must show that there is no significant loss of information as a result of this transformation. We consider the task of differentiating tweets that contain a relevant reference from those that do not from a corpus of tweets containing references to diseases.

For a tweet to be relevant it must contain a reference to an ongoing or recent (not exceeding a maximum time span of a few weeks) disease event and its location must be determinable and not aggregated beyond a users' geographical country. All other references are deemed irrelevant. To test our model we compare results obtained on the baseline against those obtained with our ontology features. In addition we investigate the impact of using N-grams and word vectors in concert with our Concept Normal Form representation. Since the numbers of positive and negative examples in these data sets are imbalanced, we opt for precision, recall and f-measure as the performance metrics. For results on the training data set we used 10 fold cross validation. We are interested in the overall performance as well as the variance in performance across datasets as we have developed our ontology so as to minimize the effect of lexical differences between documents. For each method we use a single model trained on the flu + Listeria data set to classify all datasets and take a simple mathematical, unweighted average as the overall performance across all datasets. We also employ the performance variance as a measure of the impact of lexical differences between the data sets. A small variance means the classifier's performance on different data sets is similar and that our concept model is sufficiently general.

## 6.0. Results

The results are summarized in the table below:

| Method | Metric | Influenza + Common Cold + Listeria | Conjuctivitis | Conjunctivitis 2 (Pink eye) | Norovirus | Stomach flu (Gastro Enteritis) | Overall Performance | Variance |
|---|---|---|---|---|---|---|---|---|
| **Unigram BOW + SGD (Baseline)** | F1 Score | 0.8435 | 0.6746 | 0.6607 | 0.0016 | 0.8860 | 0.6133 | 0.1269 |
| | Precision | 0.8144 | 0.8120 | 0.7020 | 0.0075 | 0.8370 | 0.6346 | 0.1256 |
| | Recall | 0.8747 | 0.5770 | 0.6240 | 0.0009 | 0.9410 | 0.6035 | 0.1379 |
| **Unigram BOW + CNF + SGD** | F1 Score | 0.7278 | 0.7592 | 0.5946 | 0.5929 | 0.8954 | 0.7140 | 0.0160 |
| | Precision | 0.7387 | 0.6599 | 0.4265 | 0.5679 | 0.8713 | 0.6529 | 0.0284 |
| | Recall | 0.7173 | 0.8937 | 0.9814 | 0.6203 | 0.9209 | 0.8267 | 0.0230 |
| **Unigrams + Bigrams + CNF + SGD** | F1 Score | 0.8130 | 0.7375 | 0.6270 | 0.5288 | 0.8766 | 0.7166 | 0.0197 |
| | Precision | 0.7890 | 0.6813 | 0.4702 | 0.6247 | 0.8976 | 0.6926 | 0.0264 |
| | Recall | 0.8385 | 0.8038 | 0.9405 | 0.4584 | 0.8566 | 0.7796 | 0.0348 |
| **Doc2Vec + Logistic regression + CNF** | F1 Score | 0.7043 | 0.8940 | 0.8489 | 0.9344 | 0.9049 | 0.8573 | 0.0083 |
| | Precision | 0.6977 | 0.9068 | 0.8194 | 1.0000 | 0.9330 | 0.8714 | 0.0136 |
| | Recall | 0.7111 | 0.8815 | 0.8806 | 0.8768 | 0.8784 | 0.8457 | 0.0057 |

*Table 1: CNF Vs Unigram Bag of Words*

The results overall indicate that better generalization is achieved with our method as the results variance is overall smaller when the CNF notation is used. For instance there is a nearly 700% reduction in model performance variance across data sets in comparison to the baseline just by transforming the data sets into the CNF representation without changing any other model parameter using unigram bag of words. The implication is that using the CNF, it is possible to create good generic classifiers for disease detection even on noisy data with high lexical diversity such as tweets.

We find there is little benefit in applying N-grams in terms of performance improvement across datasets by F1-measure. The only marked improvement occurs on the training data and in general anything beyond bigrams does not return any noticeable gains. In comparison to the unigram model there appears to be an increase in precision combined with a reduction in recall. However, the most remarkable results by far are produced when we use Doc2Vec in combination with the CNF representation. As shown in Table 1, it is the strongest approach both in terms of maximizing overall classification performance and minimizing performance variance across data sets. However, as expounded in section 5.3 it requires giving some thought to what a good word vector model would mean in terms of a conceptual representation. As a matter of fact some of our initial models returned extremely poor results although generally speaking it is our observation that word vectors improve the performance. We should also note that we have performed very little tuning beyond tuning the CNF word vectors. This was because from the onset we aimed to produce a general classifier but none the less we find the results to be remarkably good especially on the testing data sets.

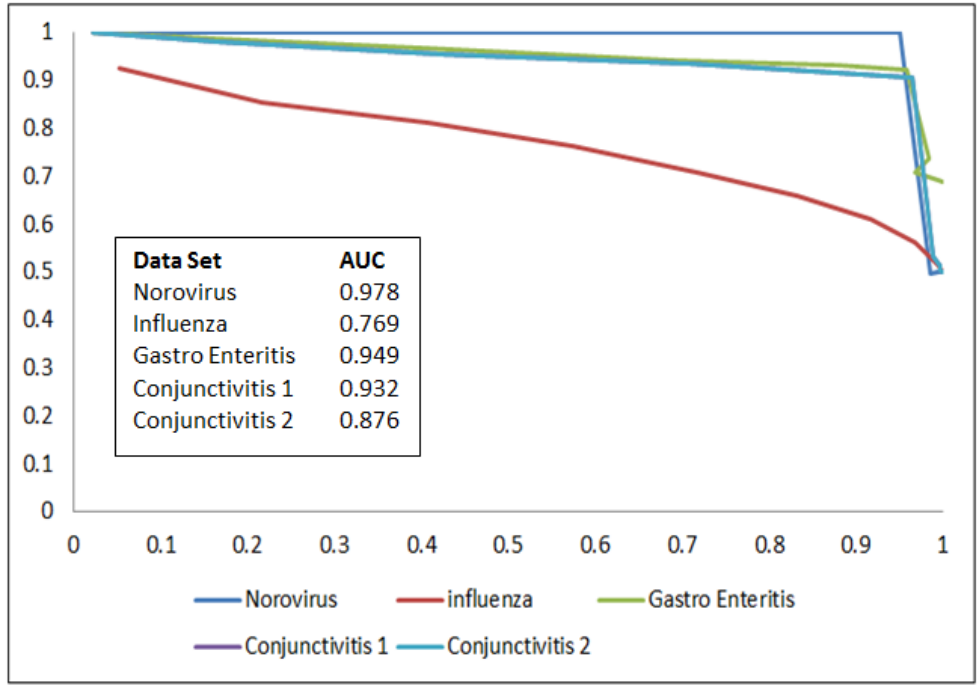

*Figure 7: Precision Recall Curve for Doc2Vec + CNF experiment*

Figure 7, shows the precision recall behavior for the model on the different data sets for the best performing Doc2Vec model at different probability thresholds. The results in table 1 are for the default threshold of 0.5. The Logistical Regression classifier actually outputs a probability. For a given threshold, any results with a higher probability of being relevant than the threshold are classified positive. The larger the area under the curve (AUC) on the Precision Recall Curve the better the classifier in terms of Precision and Recall. Our results therefore, as indicated in the summary table in the plot, are testament to the strength of the Doc2Vec approach.

## 7.0. Discussion:

In information retrieval experiments care is usually taken to ensure that the vocabulary of the training and testing data are similar, in our case we have deliberately complicated the task by using a different sets of diseases to train and test. This was to simulate what would happen if our model encountered data with a divergent lexicon as would be the case if it were deployed into an online system. The results show that our ontology driven approach is more robust than word level lookup methods like unigram bag of words.

Furthermore, the discovery that word2vec/Doc2vec vectors learned from our ontology concepts give good classification performance is of great significance. This is because although word2vec/Doc2vec embeddings have been shown to dramatically improve the performance of text classification tasks, they rely on word level look up and there is no way to compute vectors for words that do not appear in the training corpus. However, the CNF is capable of representing any document using a fixed length vocabulary that is a concatenation of our ontology concepts and part of speech information. The fixed length vocabulary implies that there can be no out of vocabulary tokens. Therefore, provided good vectors can be obtained for each concept very robust models can be obtained.

## 8.0. Conclusion:
We have described an ontology of concepts used to communicate disease incidence. We employed the ontology generate features for a classification model to partition a corpus of tweets containing mentions of diseases into relevant and irrelevant tweets. We show that our ontology based method not only allows for deeper semantic criteria to be employed. It also not only achieves high performance but results in more robust models whose performance is more stable on lexically divergent data sets.

## 9.0. Acknowledgements:

In addition I would like to thank the NORHED (Norwegian Program for Capacity building in Higher Education and Research development) for funding this research through its HI-TRAIN (Health Informatics Training and Research in East Africa for Improved Health Care) project.